\documentclass{article}
\usepackage{arxiv}
\usepackage{cite}
\usepackage[utf8]{inputenc} 
\usepackage[T1]{fontenc}    
\usepackage[hypertexnames=false]{hyperref}
\usepackage{url}            
\usepackage{booktabs}       
\usepackage{amsfonts}       
\usepackage{nicefrac}       
\usepackage{microtype}      
\providecommand{\lipsum}[1]{} 
\usepackage{graphicx}
\usepackage{float} 
\usepackage{amsmath}
\usepackage{subcaption}

\providecommand{\keywords}[1]{%
  \par\smallskip\noindent\textbf{Keywords: }#1\par\smallskip
}
\title{Geometry-Aware Sparse Depth Sampling for High-Fidelity RGB-D Depth Completion in Robotic Systems}

\author{\href{https://orcid.org/0000-0003-1072-3686}{\includegraphics[scale=0.06]{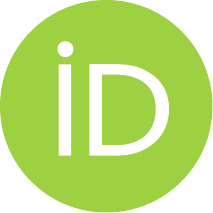}\hspace{1mm}Tony Salloom} \\
	Saisuode (Shanghai) Intelligent Technology Co., Ltd. (Synthoid.ai)\\
	Shanghai, China\\
	\texttt{Tony.salloom@synthoid.ai} \\
	\and
	Dandi Zhou \\
	Saisuode (Shanghai) Intelligent Technology Co., Ltd. (Synthoid.ai)\\
	Shanghai, China \\
	\texttt{zhou.dandi@synthoid.ai} \\
	\and
    Xinhai Sun \\
	Saisuode (Shanghai) Intelligent Technology Co., Ltd. (Synthoid.ai)\\
	Shanghai, China \\
	\texttt{sun.xinhai@synthoid.ai} \\
}

\date{}

\renewcommand{\headeright}{}
\renewcommand{\undertitle}{}

\begin{document}
\maketitle

\begin{abstract}
	Accurate three-dimensional perception is essential for modern industrial robotic systems that perform manipulation, inspection, and navigation tasks. RGB-D and stereo vision sensors are widely used for this purpose, but the depth maps they produce are often noisy, incomplete, or biased due to sensor limitations and environmental conditions. Depth completion methods aim to generate dense, reliable depth maps from RGB images and sparse depth input. However, a key limitation in current depth completion pipelines is the unrealistic generation of sparse depth: sparse pixels are typically selected uniformly at random from dense ground-truth depth, ignoring the fact that real sensors exhibit geometry-dependent and spatially nonuniform reliability. In this work, we propose a normal-guided sparse depth sampling strategy that leverages PCA-based surface normal estimation on the RGB-D point cloud to compute a per-pixel depth reliability measure. The sparse depth samples are then drawn according to this reliability distribution. We integrate this sampling method with the Marigold-DC diffusion-based depth completion model and evaluate it on NYU Depth v2 using the standard metrics. Experiments show that our geometry-aware sparse depth improves accuracy, reduces artifacts near edges and discontinuities, and produces more realistic training conditions that better reflect real sensor behavior.
\end{abstract}

\keywords{emantic space \and Point cloud \and Depth completion \and Robotic \and Sparse depth}

\section{Introduction}
\label{sec:intro}

\begin{figure}[ht]
    \centering
    \includegraphics[width=\textwidth]{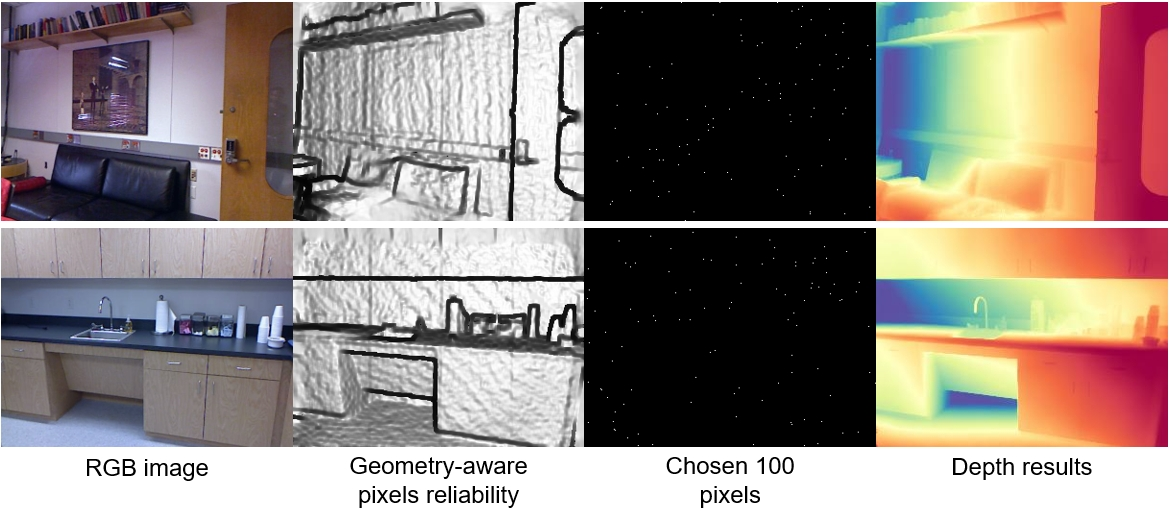}
	\caption{Applying a geometry-aware sparse depth selection method. The brighter the pixel, the higher its reliability.}
	\label{fig1:fig1}
\end{figure}

In modern industrial robotics, accurate three-dimensional scene perception is fundamental for tasks such as object recognition, manipulation, and navigation. Many robotic systems rely on RGB-D cameras—which provide synchronous color and depth streams—or stereo vision systems to perceive their environment in 3D. However, in practice, the depth information from these sensors is often imprecise, incomplete, or corrupted by noise, which significantly complicates downstream tasks.

For example, commodity RGB-D cameras often produce depth maps with large missing regions (holes) and inaccuracies, particularly when the surfaces are shiny, transparent, highly reflective, or very distant ~\cite{Zhang2018}.

Similarly, stereo-based depth detection systems exhibit depth error distributions that strongly depend  on scene geometry, texture, and baseline, and the error tends to increase with distance or in low‐texture areas ~\cite{ZhangBoult2011}.
Because of these limitations, many approaches apply a second stage — depth completion or refinement — to convert raw depth (often sparse or noisy) into a more reliable dense depth map suitable for robotics use.

In the depth-completion paradigm, the most recent deep-learning approaches take as input a color image (or images) and a (typically sparse) depth map, and learn to predict a dense, refined depth map ~\cite{Zhang2018}.

Although effective, such methods implicitly assume that the sparse depth input (or the raw depth sensor output) contains sufficient accurate measurements in roughly correct locations from which the network can propagate depth to nearby pixels. In reality, the original depth map produced by the sensor may not simply be sparse, but may be dense yet highly erroneous—with only a few pixels genuinely reliable, many others significantly biased or missing—and the spatial distribution of accurate versus inaccurate depths varies considerably with the scene structure, lighting, and sensor configuration. In other words: the “good” points are hidden in the full depth map, and there is no explicit indication of which pixels are trustworthy and which are not. Despite this, almost all depth‐completion networks treat the input depth map uniformly (or apply simple confidence maps), rather than explicitly selecting the most accurate regions for propagation.

Moreover, the input sparce depth map is created by sampling the ground truth depth for training, This makes the neural network model learn that those depth are truth points, and there is no need to refine them, thus the model fixes the depth value of sampled points and builds the depth map based on them, this sometimes amplifies the error in the area around them. In robotics contexts—especially industrial settings where safety, precision, and reliability are essential—this issue becomes particularly salient. The performance of robot perception, 3D reconstruction, and downstream manipulation is highly dependent on the fidelity of the depth map. If refinement is applied globally on regions that include grossly erroneous depths, the resulting dense map may still contain unacceptable errors or propagate the wrong geometry. Therefore, selecting only the most reliable depth samples from the raw sensor output could plausibly improve the depth‐completion process, allowing the refinement network to focus on propagation from trustworthy seeds rather than trying to salvage heavily corrupted data.
Current geometry-only sampling (based on surface normals) ignores semantic context—for example, a "metal gear" and a "plastic cover" in an industrial scene may have similar surface normals but different depth reliability (e.g., metal surfaces are more reflective and prone to depth noise for RGB-D sensors). The semantic space can encode these category-specific properties to refine the depth reliability estimation.

In this work, we address this limitation by proposing a geometry-aware sampling strategy based on local surface normals. For each pixel, we compute a PCA-based normal vector from the point cloud reconstructed from RGB and depth. Using the normal consistency and curvature within the local neighborhood, along with viewing angle considerations, we derive a depth reliability score. The sparse depth pixels are then sampled proportionally to this reliability map, yielding sparse depth maps that better mimic real-world sensor behavior.

We integrate our method with the Marigold-DC diffusion-based depth completion model and evaluate it on NYU Depth v2. Our contributions are:
\begin{itemize}
    \item A theoretical analysis of depth error distributions in RGB-D and stereo systems.
    \item A normal-guided sparse depth sampling method based on PCA-derived local geometry.
    \item Integration of this method into a diffusion-based depth completion framework.
\end{itemize}

By combining these contributions, our approach aims to improve the input to deep networks for depth refinement in robotic environments, thereby increasing the reliability of 3D perception in industrial robotics settings.

\section{Related Work}
\label{sec:related}

\subsection{Depth Sensing in Robotics}
\label{sec:rel_depth_in_industry}
RGB-D and stereo vision sensors are now standard components of robotic workcells for tasks such as object localization, bin-picking, assembly, and inspection. Recent systems have used RGB-D sequences to build semantic 3D scene representations for manipulation and navigation, for example, using scene graphs or panoptic 3D maps ~\cite{Wu2021, Yasir2022}. 
Dense RGB-D SLAM pipelines tailored for mobile and industrial robots further highlight the importance of accurate depth in dynamic indoor environments ~\cite{Canovas2021}.
Modern industrial deployments increasingly rely on higher-quality time-of-flight (ToF) and structured-light sensors, as well as multi-camera RGB-D rigs. New datasets and platforms such as ToF-360, which provides omnidirectional ToF RGB-D scans, are explicitly designed for robotics and single-capture 3D perception ~\cite{Kanayama2025}.
Noise-aware reconstruction pipelines using multiple robot-mounted RGB-D cameras likewise emphasize that raw depth is far from perfect and must be modeled and filtered before being used for large-scale mapping or inspection ~\cite{Yang2020, Maken2025}.

At the same time, application-oriented studies on semantic RGB-D perception for robots, 3D instance segmentation, and industrial automation point out that many higher-level tasks (pose estimation, grasp planning, collision avoidance) are critically sensitive to depth quality ~\cite{Yasir2022, DeTone2018}.
These works collectively motivate more realistic modeling of sensor behavior and learning-based methods that are robust to the non-ideal characteristics of industrial depth sensors.

\subsection{Depth Completion and Sparse-to-Dense Methods}
\label{sec:rel_depth_compl_method}
Depth completion—predicting dense depth from sparse measurements and an RGB image—has been intensively studied in recent years, with several surveys synthesizing classical and deep methods ~\cite{Hu2022, Kha2022}.
Early deep approaches focused on LiDAR-to-image completion for autonomous driving, while recent methods target broader sensor setups (LiDAR, ToF, RGB-D) and extreme sparsity regimes.

Image-guided depth completion has evolved from convolutional encoder–decoders to architectures that explicitly model long-range context and cross-modal interactions. Transformers (GuideFormer), repetitive image-guided networks (RigNet), and multi-modal characteristic guided networks all exploit correlations between RGB structure and sparse depth cues ~\cite{Lee2023, Yan2025a, Yan2021, ElYabroudi2022, Rho2022}.
Propagation-based models, such as bilateral or masked spatial propagation networks, iteratively refine initial depth predictions to better respect edges and adapt to varying sparsity levels ~\cite{Tang2024, Jun2024}.
Recent work on flexible or sparsity-adaptive depth completion explicitly considers varying sampling densities and non-uniform point distributions ~\cite{Long2024, Park2024}.
In parallel, there is growing interest in multimodal and unsupervised depth completion. Efficient multimodal models integrate sparse ToF or LiDAR depth with RGB while maintaining low computational cost ~\cite{Xing2025, Hou2022}, and unsupervised methods leverage photometric consistency or distribution matching between real and simulated depth to remove the need for dense ground truth ~\cite{Gou2025, Yang2025}. Specialized approaches address challenging phenomena such as transparent objects through local implicit functions on RGB-D input ~\cite{Zhu2021}.

More recently, diffusion-based and continual-learning methods have reshaped the problem formulation. Marigold-DC and related works repurpose diffusion models trained for monocular depth estimation, treating depth completion as image-conditional generation guided by sparse depth ~\cite{Ke2025, Viola2025}. DenseFormer similarly integrates diffusion mechanisms into depth completion ~\cite{Yuan2025}.
ProtoDepth introduces a prototype-based framework for unsupervised continual depth completion, emphasizing robustness to distribution shifts and domain evolution ~\cite{Rim2025}.
Community challenges, such as the MIPI RGB+ToF depth completion challenge, provide standardized datasets and comparisons that highlight the strengths and weaknesses of current architectures and training strategies ~\cite{Zhu2023}.

All these works, however, typically assume that the sparse depth used during training is obtained by randomly sub-sampling ground truth depth, which is a convenient but unrealistic approximation of how real sensors behave.

\subsection{Limitations of Random Sparse Depth Sampling}
\label{sec:rel_Liit_random_spars}
Most supervised depth completion pipelines construct training pairs by randomly sampling a fixed number of pixels from dense ground-truth depth maps, both in indoor and autonomous driving benchmarks. This practice is visible across many contemporary methods, including transformer-based, propagation-based, multimodal and diffusion-based architectures ~\cite{Rho2022, Tang2024, Park2024, Yuan2025, Viola2025}.
While random sampling is simple and reproducible, it implicitly assumes that: (i) every pixel is equally likely to be measured by the sensor, and (ii) the retained depths are noise-free, since they come directly from ground truth.

Several recent works have begun to question this assumption. Sparse-to-Dense Depth Completion Revisited systematically studies the effect of sampling patterns and shows that Poisson-disk sampling leads to better performance than uniform random sampling, because it enforces a more realistic spatial coverage of the scene ~\cite{Xiong2020}. SparseDC and Flexible Depth Completion further highlight that real sensors produce highly non-uniform and scene-dependent sampling patterns, and that methods trained only on fixed, random sparsity struggle when evaluated under different densities or distributions ~\cite{Park2024, Long2024}.
Beyond static sampling, Uncertainty-Aware Interactive LiDAR Sampling proposes actively selecting LiDAR rays based on uncertainty estimates from a depth completion network, effectively biasing measurements toward informative regions of the scene ~\cite{Taguchi2023}.
Similarly, sparsity-adaptive propagation networks adjust their refinement behavior according to the actual number and layout of sparse depth points, implicitly acknowledging that real-world sampling is not random and not fixed ~\cite{Jun2024, Rim2025}.

Despite these advances, the majority of works still treat the available sparse depth values as perfect observations drawn from a uniform distribution over pixels. This stands in contrast to empirical studies of depth sensors, which show that measurement reliability strongly depends on local surface orientation, distance, material, and occlusion ~\cite{Maken2025, Mehltretter2022}.
 In particular, planar regions at favorable viewing angles tend to yield more accurate depth, while edges, grazing angles, reflective surfaces, and occlusions produce outliers, missing values, or heavy-tailed noise.

Our work addresses this gap by proposing a sparse-depth selection strategy that is explicitly conditioned on local surface geometry, as captured by PCA-based normals on the point cloud. Rather than treating all pixels as equally reliable, we use geometric cues to mimic the heterogeneous reliability patterns observed in real RGB-D and stereo systems. This provides a more realistic training signal for depth completion networks, including diffusion-based models such as Marigold-DC, and better aligns the synthetic sparse depth used in training with the structured error distributions observed in practice.

\section{Research methodology}
\label{sec:methodology}
In this research paper, we devides our research into two steps. i) First, we analyze the source of the depth error in RGBD and stereo camera depth maps. Since we do not have nor can we buy all type of cameras to do emperical analyzes, we depend on the previous research in this fields and do a brief survey about the source of error in the depth maps provided by cameras.
ii) second, we introduce a sparce selection method, which has been designed based on our finding from previous step that depends on our findings in the first step. Then we choose the Marigold ~\cite{Viola2025} depth completion model and the NYU depth dataset version 2 ~\cite{Wang2022} to experiment with our algorithm and analyze the results.

\section{Sources of Error in RGB-D and Stereo Depth Estimation}
\label{sec:rel_error_in_depth}
A line of recent work explicitly models the noise characteristics of RGB-D sensors deployed on robots. For structured-light RGB-D cameras such as Zivid, depth noise has been shown to depend on distance, incidence angle, ambient illumination, exposure time, and multi-frame acquisition settings; accurate noise models substantially improve downstream 3D reconstruction quality ~\cite{Maken2025}.
For ToF sensors, detailed studies analyze systematic and random errors due to multipath interference, shot noise, non-linearities and low resolution, and propose compensation schemes that depend on scene geometry and sensor configuration ~\cite{Wang2022, Rodriguez2023}.
These findings are reflected in new RGB-D datasets and benchmarks: for example, ToF-centric datasets highlight stitching artifacts, edge distortion, and anisotropic noise that arise from panoramic scanning and multi-path effects ~\cite{Yang2020, Kanayama2025}.
Such characteristics lead to depth maps whose error distribution varies across pixels and surfaces, contradicting the simplified assumption of independent, identically distributed noise.

In stereo depth estimation, recent work focuses on quantifying and learning uncertainty. Evidential deep stereo matching and joint depth–uncertainty estimation explicitly model aleatoric and epistemic uncertainty, providing pixel-wise confidence alongside disparity ~\cite{Jeong2025, Wang2022}.
New training schemes weight the loss by uncertainty to encourage the network to focus on hard regions such as occlusions and weak texture ~\cite{Jeong2025, Cheng2025}.

These studies make two key points relevant to our work: (1) error statistics depend strongly on local geometry, occlusion, and texture, and (2) confidence or reliability varies spatially, even within a single frame.

Overall, both RGB-D and stereo literature converge on the view that the distribution of depth error is spatially heterogeneous and structured by surface orientation, material, incidence angle, and occlusion—precisely the aspects that our normal-based sparse-depth selection aims to exploit.

\section{Proposed Method: Normal-Based Sparse Depth Selection}
\label{sec:Proposed Method}
\begin{figure}
	\centering
    \includegraphics[width=\textwidth]{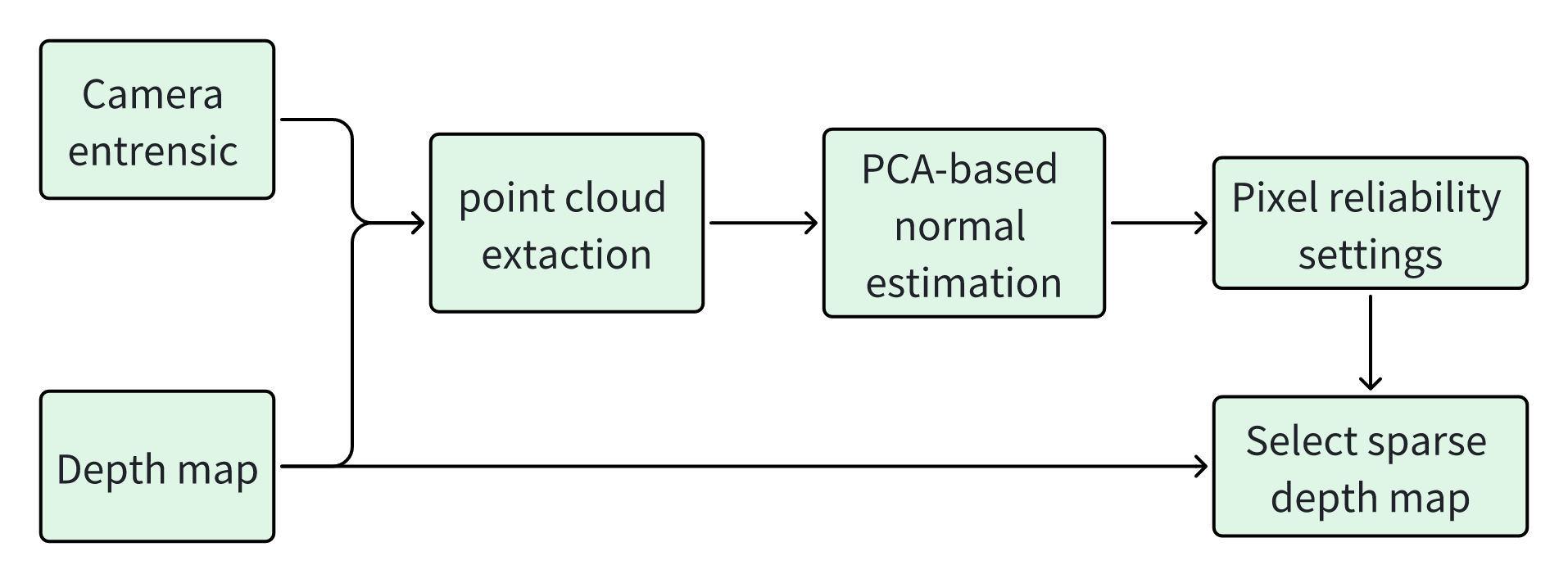}
	\caption{Sparse depth creation algorithm flowchart.}
	\label{fig1:proccess}
\end{figure}

We propose a geometry-aware sparse depth sampling strategy that better reflects the heterogeneous reliability of real depth sensors. Instead of selecting sparse depth pixels uniformly at random from a ground-truth depth map, we estimate local surface normals and planarity from the corresponding point cloud and use them to define a depth reliability score for each pixel. Sparse depth points are then drawn according to this reliability distribution and passed, together with the RGB image, to a depth completion network (in our experiments, Marigold-DC). The process flow chart is shown in fig. \ref{fig1:proccess}

\subsection{Point Cloud Extraction}
\label{sec:point extraction}
The point cloud of the scene is extracted based on the intrensic parameters of the camera and the ground truth depth map using the known projection method.
For each pixel ($u,v$) (zero-based indexing), the corresponding 3D point in camera coordinates is obtained by back-projection:
\begin{equation}
\label{eq:pnt from depth}
    \begin{bmatrix}
        X_{(u,v)} \\
        Y_{(u,v)} \\
        Z_{(u,v)} \\
    \end{bmatrix} = D(u,v) K^{-1} \begin{bmatrix}
        u \\
        v \\
        1
    \end{bmatrix}
\end{equation}
where $D\in \mathcal{R}^{H\times W}$, its corresponding ground-truth depth map and $K$ is the camera intrensic parameters.

\subsection{Surface Normal Estimation via PCA}
\label{sec:normal est pca}
For each point $p_i\in \mathcal{P}$, we estimate a local surface normal using Principal Component Analysis (PCA) over its spatial neighborhood.

Let $mathcal{N}(i)$ denote the index set of neighboring points of $p_i$ (e.g., a fixed-radius 3D neighborhood or a $k \times k$ pixel window in the image). The local mean is
\begin{equation}
    \boldsymbol{\mu}_i 
    = \frac{1}{|\mathcal{N}(i)|} 
    \sum_{j \in \mathcal{N}(i)} \mathbf{p}_j
\end{equation}
The corresponding covariance matrix is
\begin{equation}
    \mathbf{C}_i 
    = \frac{1}{|\mathcal{N}(i)|}
    \sum_{j \in \mathcal{N}(i)}
    \left( \mathbf{p}_j - \boldsymbol{\mu}_i \right)
    \left( \mathbf{p}_j - \boldsymbol{\mu}_i \right)^\top
\end{equation}
We perform eigen-decomposition
\begin{equation}
    \mathbf{C}_i \mathbf{v}_{i,k} = \lambda_{i,k} \mathbf{v}_{i,k}, \qquad k \in \{1,2,3\}
\end{equation}
with eigenvalues ordered as
\begin{equation}
    \lambda_{i,1} \le \lambda_{i,2} \le \lambda_{i,3}
\end{equation}
The eigenvector associated with the smallest eigenvalue, $v_{i,1}$, corresponds to the direction of least variance and is taken as the surface normal $p_i$:
\begin{equation}
    \mathbf{n}_i = \mathbf{v}_{i,1}, \qquad \|\mathbf{n}_i\|_2 = 1
\end{equation}
the local surface curvature can be approximated by the normalized smallest eigenvalue
\begin{equation}
    \kappa_i 
    = \frac{\lambda_{i,1}}{\lambda_{i,1} + \lambda_{i,2} + \lambda_{i,3}}
\end{equation}
Small $\kappa_i$ indicates a locally planar patch; large $\kappa_i$ suggests edges or corners. Optionally, we orient normals to face the camera by enforcing
\begin{equation}
    \mathbf{n}_i \leftarrow 
    \begin{cases}
        \mathbf{n}_i, & \mathbf{n}_i^\top \mathbf{p}_i < 0 \\
        -\mathbf{n}_i, & \text{otherwise}
    \end{cases}
\end{equation}

\subsection{Depth Reliability Metric and Sampling Strategy}
\label{sec:reliability}
We define a depth reliability score $r_i \in [0,1]$ for each point $p_i$ based on the angle between the normal at that point and the vector from the point to the camera position, which is the coordinate origin.

Let $v_i$ be the viewing direction from the camera center to $p_i$:
\begin{equation}
    \mathbf{v}_i = \frac{\mathbf{p}_i}{\|\mathbf{p}_i\|_2}
\end{equation}
The angle of incidence $\theta_i$ between the normal and the viewing direction is
\begin{equation}
    \cos \theta_i = |\mathbf{n}_i^\top \mathbf{v}_i|
\end{equation}
Depth measurements are generally more reliable for moderate incidence angles and degrade as $\theta_i$ approaches $90$. We define
\begin{equation}
    s_{\text{angle}, i} = \cos^{\beta} \theta_i
\end{equation}
with $\beta \geq 1 $ shaping the penalty for grazing angles.
The final reliability score is given by $ r_i = s_{\text{angle},i}$. 

To obtain a sparse depth map with a target number of points $K$, we convert the reliability scores into sampling probabilities with the following equation. 
\begin{equation}
    p_i = \frac{r_i}{\sum_{j=1}^{N} r_j}
\end{equation}
We then draw $K$ indices $\{ i_1, \ldots, i_K \}$ without replacement according to $p_i$. The resulting sparse depth map $D_{sp}$ is defined as:
\begin{equation}
    D_{\text{sp}}(u,v) =
    \begin{cases}
        D(u,v), & \text{if } (u,v) \text{ corresponds to a sampled index } i_k\\
        0, & \text{otherwise.}
    \end{cases}
\end{equation}
\subsection{Integration with the Marigold-DC Model}
\label{sec:integrating}
We integrate our sampling strategy into the Marigold-DC depth completion framework without altering the model architecture. Marigold-DC is a diffusion-based depth completion system that conditions a pre-trained monocular depth diffusion model on sparse depth samples and RGB image features.

During training, we replace the original random sparse-depth sampling step with our normal-guided sampling method, producing a sparse depth map that more closely reflects realistic sensor behavior. The RGB image, sparse depth, and all encoder/decoder components of Marigold-DC remain unchanged. Training is performed using the same hyperparameters, augmentation strategy, and optimization schedule reported in the Marigold-DC paper to ensure comparability.

At inference time, Marigold-DC receives the sparse depth points generated by our method and produces a dense depth prediction through its diffusion refinement process.

\section{Experimental Setup}
\label{sec:experement setup}
\subsection{Dataset}
\label{sec:dataset}
We conducted all experiments on the NYU Depth v2 indoor RGB-D dataset, following the same evaluation protocol used in the original Marigold-DC paper. The dataset consists of over 400K RGB-depth pairs acquired with a Microsoft Kinect structured-light camera in variouse indoor environments (living rooms, offices, kitchens, and classrooms). However, only 500 frames are used for evaluating the proposed selection method. We used the Marigold-dc pretrained model for evaluation. In this paper, no training is neede. 
\subsection{Integration with the Marigold-DC Model}
We integrate our sampling strategy into the Marigold-DC depth completion framework without altering the model architecture. Marigold-DC is a diffusion-based depth completion system that conditions a pre-trained monocular depth diffusion model on sparse depth samples and RGB image features.
For normal estemation with PCA, we set the neighboring area size to $5\times 5$mm, with minimum number of points = 5. 

At inference time, Marigold-DC receives the sparse depth points generated by our method and produces a dense depth prediction through its diffusion refinement process.

\subsection{Baselines for Comparison}
To assess the effectiveness of our proposed sparse-depth sampling strategy, we compare against the Marigold original random sampling method, where the probability of selecting each depth pixel is  uniformly distributed.

All baselines use the same number of sparse depth points as our method for fair comparison.  The size of sparse depth is set to 100, 200, 300, and 500.

We evaluate all methods in the NYU Depth v2 test set using the same quantitative metrics reported in the Marigold-DC paper ~\cite{Viola2025}, including root mean squared error (RMSE) and the mean absolute error (MAE)

\subsection{Results and Discussion}
\label{sec:results}
This section presents quantitative and qualitative results of our geometry-aware sparse depth sampling strategy, comparing it against the uniformly random sampling baseline for depth completion.
\subsubsection{Quantitative Results}
Table \ref{table1:res} ummarizes the depth completion performance (measured by Mean Absolute Error, MAE, and Root Mean Squared Error, RMSE) of our geometry-aware sampling and the uniformly random sampling method across different numbers of sparse pixels.
For all sparse pixel counts (100–500), our geometry-aware strategy outperforms random sampling:
\begin{itemize}
    \item At 100 sparse pixels, our method reduces RMSE by 12.0\% (from 0.211 to 0.186) and MAE by 3.4\% (from 0.089 to 0.086).
    \item Even at 500 sparse pixels (near-dense sampling), our method still achieves a 5.7\% RMSE reduction (0.105 → 0.099), with MAE matched (0.039).
    \item The largest performance gap occurs at 300 sparse pixels: RMSE drops by 7.5\% (0.159 → 0.147) and MAE by 6.5\% (0.062 → 0.058), highlighting the value of geometry-aware sampling for mid-density sparse inputs.
\end{itemize}

\begin{table}[ht]
\centering
\caption{Comparison results between our geometry-aware selection method and the uniformly random selection.}
\begin{tabular}{c|cc|cc}
\hline
 & \multicolumn{2}{c|}{\textbf{Geometry-Aware sampling}} 
 & \multicolumn{2}{c}{\textbf{Uniformly random sampling}} \\
\textbf{Number of sparse pixels} & MAE & RMSE & MAE & RMSE \\
\hline
100 & 0.086 & 0.186 & 0.089 & 0.211 \\
200 & 0.079 & 0.189 & 0.086 & 0.207 \\
300 & 0.058 & 0.147 & 0.062 & 0.159 \\
500 & 0.039 & 0.099 & 0.039 & 0.105 \\
\hline
\end{tabular}
\label{table1:res}
\end{table}

These results confirm that prioritizing reliable depth pixels (via PCA-based surface normals) yields more informative sparse inputs for depth completion, especially when sparse pixel budgets are limited.

\subsubsection{Qualitative Results}
\label{sec:qualitative}
\begin{figure}
    \centering
    \includegraphics[width=\textwidth]{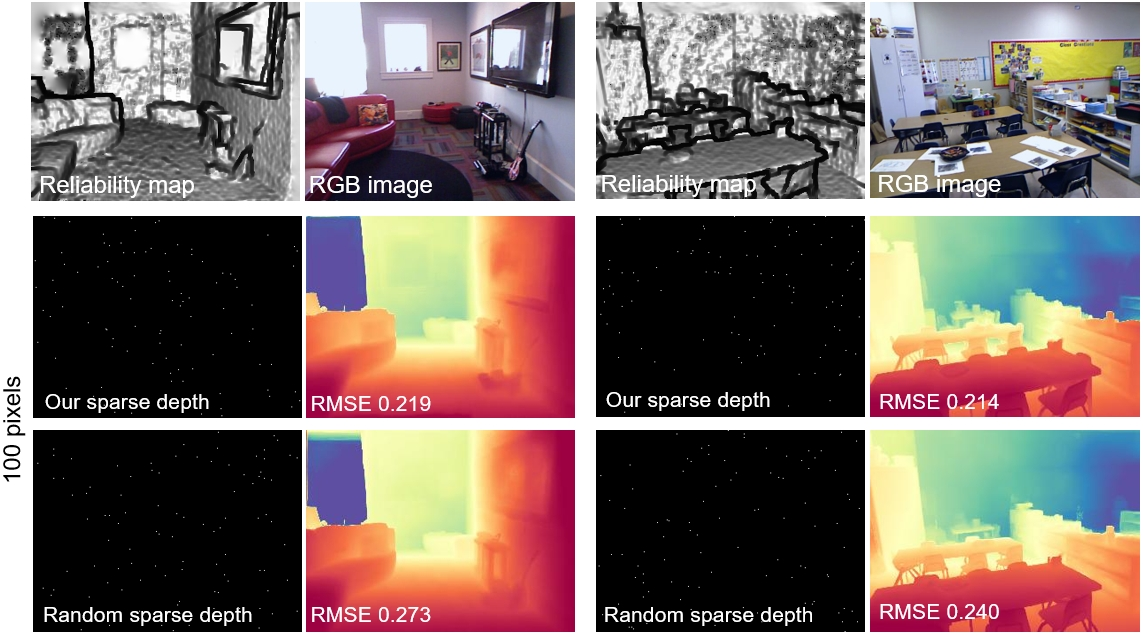}
    \includegraphics[width=\textwidth]{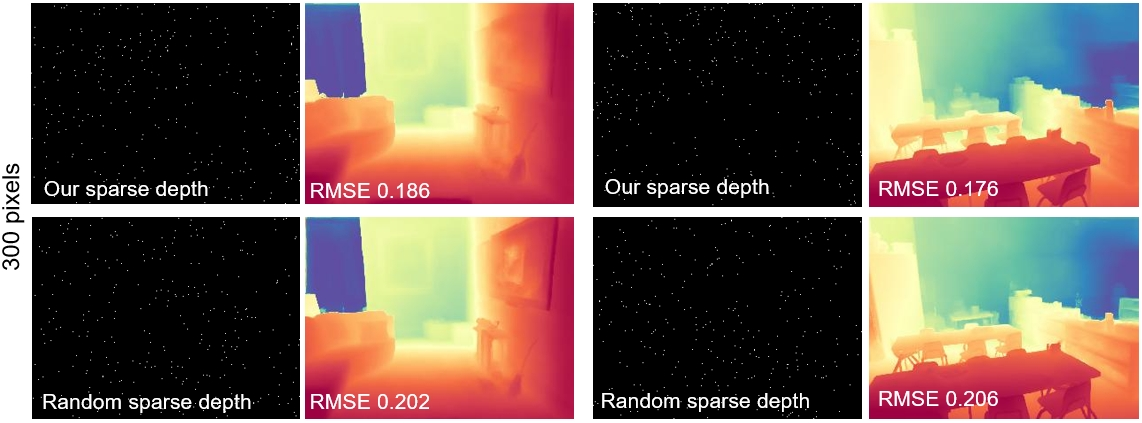}
    \includegraphics[width=\textwidth]{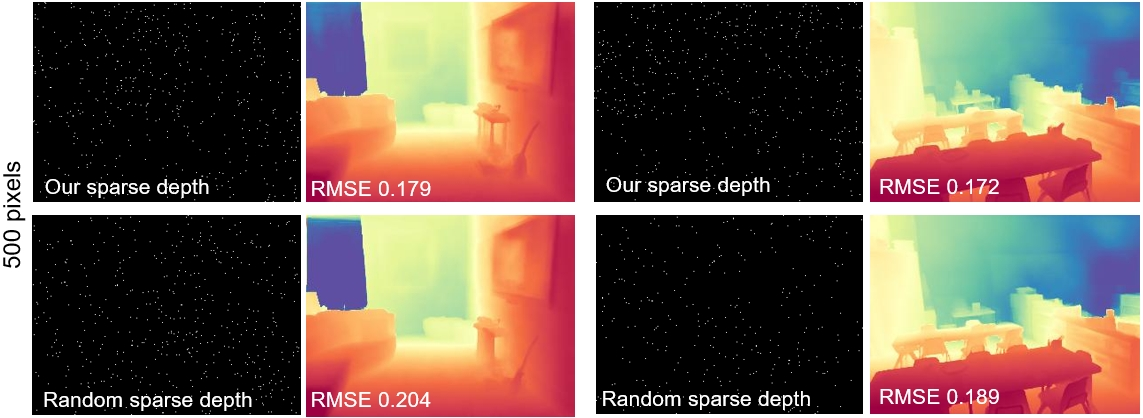}
    \caption{The imapact of different sparse depth map size on depth completion results when using our geometry-aware depth selection and the uniform random selection method.}
    \label{fig:res}
\end{figure}

Figure \ref{fig:res} visualizes qualitative depth completion results for two scene examples (left: indoor living space; right: classroom), across 100, 300, and 500 sparse pixels. Each row pair compares our geometry-aware sampling (top) with random sampling (bottom) for the same sparse pixel count.
Key observations from the figure:
\begin{enumerate}
    \item \textit{Reliability Map Alignment}: The top row of each scene shows our reliability map (sparser, focused on geometrically stable regions like flat surfaces/object interiors), while random sampling (implied by uniform pixel distribution) lacks this focus.
    \item \textit{Edge/Discontinuity Artifacts}: For both scenes (e.g., the sofa edge in the living space, desk edges in the classroom), our method produces smoother depth transitions at boundaries. Random sampling introduces more noise and misaligned edges (visible in the RMSE heatmaps as brighter, higher-error regions).
    \item \textit{RMSE Heatmap Improvement}: Across all sparse pixel counts, our method’s RMSE heatmaps (e.g., 100 pixels: 0.219 vs. 0.273 for random) show cooler (lower-error) regions, especially on object surfaces and scene structures critical for robotic perception.
\end{enumerate}

Notably, at 300 sparse pixels, our method’s RMSE (0.186 for the living space, 0.176 for the classroom) is substantially lower than random sampling (0.202 and 0.208, respectively)—consistent with the quantitative trends in Table \ref{table1:res}.

\subsubsection{Discussion}
The quantitative and qualitative results collectively demonstrate that our geometry-aware sparse sampling better mimics real sensor behavior (which prioritizes reliable geometric regions) than uniform random sampling. This leads to:
\begin{itemize}
    \item Higher depth completion accuracy (lower MAE/RMSE) across sparse pixel budgets.
    \item Reduced artifacts at scene edges and discontinuities—critical for industrial robotics tasks (e.g., manipulating objects or navigating cluttered environments).
    \item Improved generalization to realistic sensor constraints, as our sampling strategy encodes geometric reliability (absent in random sampling).
\end{itemize}

A limitation of the current framework is its focus on geometric cues alone; future work could integrate semantic space (e.g., semantic class priors for depth reliability) to further refine sampling in semantically diverse industrial scenes.
\section{Conclusion and Future Work}
\label{sec:conclusion}
We present a geometry-aware sparse depth sampling approach that employs PCA-based surface normal estimation to generate realistic sparse input data for depth completion networks. In contrast to uniform random sub-sampling, our method emulates the structured reliability patterns inherent to real-world RGB-D and stereo depth sensors. When integrated into the Marigold-DC framework, this strategy yields enhanced depth completion performance on the NYU Depth v2 dataset, as validated by both quantitative metrics and qualitative assessment.

For future research, we outline several directions: (1) end-to-end training of depth completion models using the proposed sampling strategy, coupled with rigorous evaluation of the resultant outputs; (2) incorporating a distance-dependent term into the depth reliability calculation to account for sensor performance degradation with increasing range; (3) extending the method’s compatibility to a broader spectrum of depth completion architectures; (4) embedding uncertainty quantification directly into the sparse sampling pipeline to further refine input informativeness; and (5) generating sparse depth samples via a fused scoring mechanism that combines geometric reliability (from surface normals) and semantic space reliability (from scene-level semantic priors).

\bibliographystyle{unsrt} 

\end{document}